\documentclass[10pt,twocolumn,letterpaper]{article}

\usepackage{cvpr}      
\usepackage{multirow}
\usepackage{graphicx} 
\usepackage{booktabs}

\usepackage[]{pifont}
\usepackage[dvipsnames]{xcolor}

\definecolor{cvprblue}{rgb}{0.21,0.49,0.74}
\usepackage[pagebackref,breaklinks,colorlinks,citecolor=cvprblue]{hyperref}

\title{ProS: Prompting-to-simulate Generalized knowledge for Universal Cross-Domain Retrieval}

\author{Kaipeng Fang\textsuperscript{1}\hspace{-0.2cm}\and
Jingkuan Song\textsuperscript{1}\and
Lianli Gao\textsuperscript{1}\hspace{-0.2cm}\and
Pengpeng Zeng\textsuperscript{1}\hspace{-0.2cm}\and
Zhi-Qi Cheng\textsuperscript{2}\hspace{-0.2cm} \and 
Xiyao Li\textsuperscript{3}\hspace{-0.2cm}\and
Heng Tao Shen\textsuperscript{1}\hspace{-0.2cm}\and
\textsuperscript{1} University of Electronic Science and Technology of China (UESTC)\\
\textsuperscript{2}Carnegie Mellon University\\
\textsuperscript{3}Kuaishou Technology
}

\begin{document}
\maketitle
\begin{abstract}

The goal of Universal Cross-Domain Retrieval (UCDR) is to achieve robust performance in generalized test scenarios, wherein data may belong to strictly unknown domains and categories during training. Recently, pre-trained models with prompt tuning have shown strong generalization capabilities and attained noteworthy achievements in various downstream tasks, such as few-shot learning and video-text retrieval. However, applying them directly to UCDR may not be sufficient to handle both domain shift (i.e., adapting to unfamiliar domains) and semantic shift (i.e., transferring to unknown categories). To this end, we propose \textbf{Pro}mpting-to-\textbf{S}imulate (ProS), the first method to apply prompt tuning for UCDR. ProS employs a two-step process to simulate Content-aware Dynamic Prompts (CaDP) which can impact models to produce generalized features for UCDR. Concretely, in Prompt Units Learning stage, we introduce two Prompt Units to individually capture domain and semantic knowledge in a mask-and-align way. Then, in Context-aware Simulator Learning stage, we train a Content-aware Prompt Simulator under a simulated test scenario to produce the corresponding CaDP.  Extensive experiments conducted on three benchmark datasets show that our method achieves new state-of-the-art performance without bringing excessive parameters.  Our method is publicly available at \url{https://github.com/fangkaipeng/ProS}

\end{abstract}    
\section{Introduction}
\label{sec:intro}

\begin{figure}
    \centering
    \includegraphics[width=1.0\linewidth]{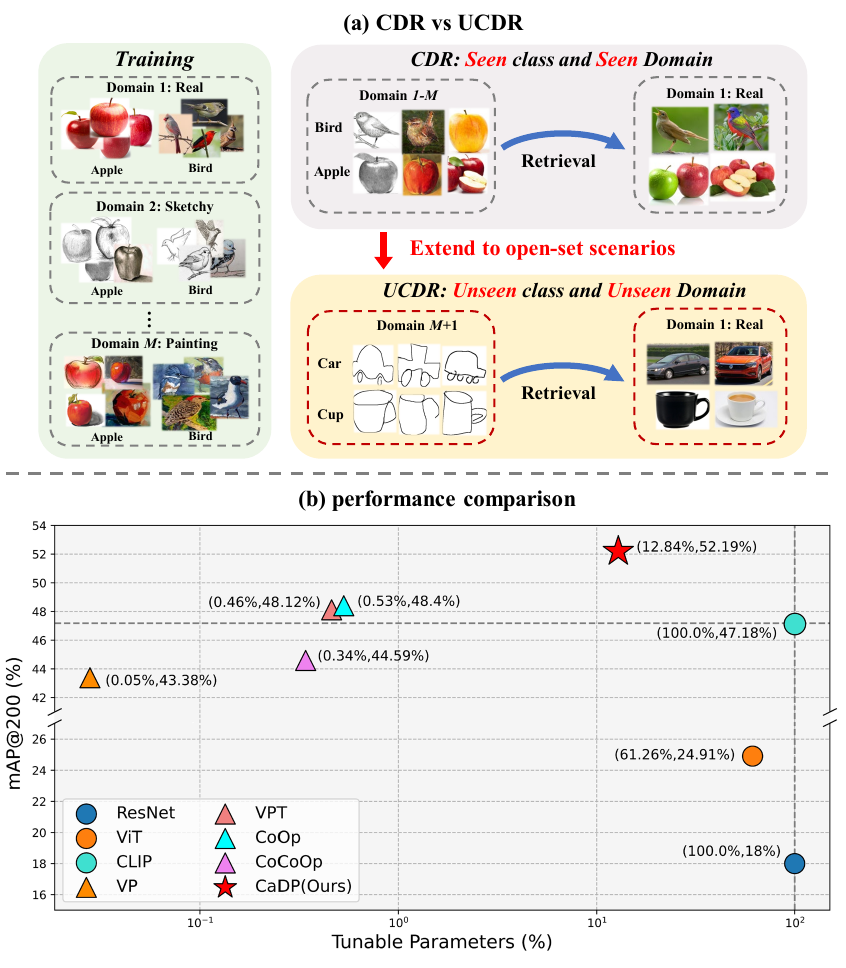}
    \vspace{-1.5em}
    \caption{\textbf{(a)} Illustration of Cross-Domain Retrieval (CDR) and its generalized version (UCDR). \textbf{(b)} Comparison of our ProS 
    \scalebox{1.5}{\ding{73}}
       with different backbones $\bigcirc$ and various prompt-based methods $\triangle$ under UCDR protocol. All prompt-based methods use CLIP as the backbone. Our method yields solid improvement and achieves a better trade-off between performance and trainable parameters usage against state-of-the art.}
    \label{fig:fig1}
    \vspace{-1.5em}
\end{figure}

Cross-Domain Retrieval (CDR)~\cite{DBLP:journals/ijmir/ZhouHLWL22,DBLP:conf/cvpr/GajicB18} is an important task in Information Retrieval (IR) systems that utilizes data from one domain (\eg Infograph, Sketch) as a query to search semantically similar examples in another domain (\eg natural image). Despite achieving promising results, existing methods~\cite{DBLP:conf/cvpr/SainBKCCXS23,DBLP:conf/cvpr/GajicB18,DBLP:conf/cvpr/BhuniaCSYXS21,DBLP:conf/cvpr/ChaudhuriBS023} heavily follow a close-set learning setting, where domains and categories in both training and testing data are pre-defined. Such evaluation limits the model to the training samples and unable to tackle the open-set applications such as e-commerce search~\cite{DBLP:conf/cvpr/ChopraSGSAK19,DBLP:conf/cvpr/FuentesS21,DBLP:conf/cvpr/LiJJLCLZZ23} and recommendation~\cite{DBLP:conf/cvpr/KangKLRM19,DBLP:conf/cvpr/FuentesS21}, where a searched product not only comes from various new domains (\eg styles) but is also accompanied by the emergence of new categories. Towards this goal, one emerging research direction is proposed, namely Universal Cross-Domain Retrieval (UCDR)~\cite{UCDR}, shown in Fig.~\ref{fig:fig1}(a), for performing well under generalized test scenarios, where test data may belong to strictly unknown domains and categories.

The core challenge for UCDR is how to empower models with the capabilities of both domain shift (\ie adapting to unfamiliar domains) and semantic shift (\ie transferring to unknown categories). The mainstream approaches~\cite{UCDR, SASA} focus on fine-tuning visual models, \eg ResNet~\cite{resnet} and ViT~\cite{ViT}, with the guidance of expert knowledge from semantic information or pre-trained models.  Compared with ResNet, ViT has stronger representational capability and contains more knowledge gained from larger pre-training data, which facilitates the generalization of the model for unknown images. This raises an intuitive question: \textit{can the extensive general knowledge inherent in large-scale pre-trained models enhance its ability to tackle the UCDR task where prior information about test samples is strictly unknown?}

With this question in mind, we empirically conduct a confirmatory experiment on the widely-recognized large-scale pre-trained model, CLIP~\cite{CLIP} under UCDR.  Specifically, we fully fine-tune the CLIP and compare its performance with ViT and ResNet, as shown in Fig.~\ref{fig:fig1} (b). From the figure, we can see that CLIP yields tremendous gains, reaching a 22.27\% improvement in mAP@200 compared with ViT. This result proves the effectiveness of common knowledge inherent in large-scale pre-trained CLIP for UCDR task.

However, the full fine-tuning strategy inevitably forgets the useful knowledge gained in the large-scale pre-training phase~\cite{Fine-Tuning_Distort_Features}. Besides, since the entire model is updated, this strategy brings significant computational costs. To alleviate above problems, we take the first step to leverage prompt tuning~\cite{coco, vpt} which is a more flexible and lightweight strategy than full fine-tuning. However, when directly applying existing prompt tuning methods~\cite{vp123,vpt,coco,cocoop} to UCDR, \ie $\triangle$ in~\cref{fig:fig1} (b), we find that these methods show a fine benefit in terms of computational cost without a strong advantage or even worse over full fine-tuning CLIP in terms of mAP@200. This may be attributed that these methods do not fulfill the properties of UCDR, contributing to poor generalization. Therefore, we pose a question:  \textit{how to design a more effective prompt tuning method for the UCDR task? }

Towards above question, we propose a novel prompt tuning method in a simulated way, named \textbf{Pro}mpting-to-\textbf{S}imulate (ProS), to effectively leverage prompt tuning to mine the generalized knowledge from CLIP for UCDR. Specifically, ProS introduces a two-stage learning to simulate two content-aware dynamic prompts (CaDP) which reflect the input sample's domain and category, respectively. In Prompt Unit Learning (PUL), two groups of distinct learnable prompts are established: domain prompt units and semantic prompt units. These two units are employed to extract domain knowledge and semantic knowledge from source data in a mask-and-align strategy. Then,  in Context-aware Simulator learning (CSL), we train a Content-aware Prompt Simulator (CaPS) under simulated test scenarios to dynamically generate content-aware dynamic prompts (CaDP) based on the learned prompt units.  With the CaDP, we can impact CLIP's knowledge to gain a more generalizable representation and achieve better performance compared with existing prompt tuning methods in UCDR, as shown in \cref{fig:fig1} (b). In addition, our method uses considerably fewer learnable parameters compared to full fine-tuning, accounting for only 12.84\% of the model parameters, and does not introduce excessive learnable parameters compared to prompt tuning methods.

Our main contributions can be summed as follows:
\begin{itemize}
    \item We are the first to investigate how to adapt CLIP with prompts for UCDR. 
    \item We propose a prompt-based method named Prompting-to-Simulate (ProS), which learns the generalized knowledge to deal with open-set scenarios. 
    \item Extensive experiments on three benchmark datasets show that our ProS achieves new state-of-the-art results compared with prompt-based methods without bringing excessive parameters.
\end{itemize}

\section{Related Work}

\subsection{Universal Cross-Domain Retrieval}

UCDR requires solving domain and category generalization simultaneously which can be roughly regarded as a combination of Domain Generalization~(DG)~\cite{DBLP:conf/iccv/LiZYLSH19, DBLP:conf/mm/WangY0W022} and Zero-Shot Learning~(ZSL)~\cite{DBLP:journals/pami/XianLSA19, DBLP:conf/mm/GeXML022}. In DG, some approaches synthesize unseen samples using data augmentation strategy~\cite{DBLP:conf/eccv/ManciniARC20} or generative adversarial networks (GANs)~\cite{DBLP:conf/eccv/ZhouYHX20} to handle the domain shift. In ZSL, some researchers leverage the auxiliary data, including human-annotated attribute information ~\cite{ZSL1, ZSL2}, text description\cite{DBLP:conf/cvpr/ReedALS16} or knowledge graph~\cite{DBLP:conf/cvpr/LeeFYW18} to learn the relationship between seen categories and pre-define unseen categories names. It should be clarified that UCDR does not require pre-defined unseen class names. Inspired by above approaches, SnMpNet~\cite{UCDR} trains the model with the mix-up data augmentation strategy and align features with semantic information. Additionally, SASA\cite{SASA} deploys a ViT-based model and trains categorical prototypes to handle the semantic shift. Different from these works, we make the first attempt to apply a large-scale pre-trained model, \ie CLIP, to the UCDR task, and propose a new prompt tuning method to improve the generalization ability of the CLIP.

\subsection{Vision-Language Pre-training Models}

Motivated by the success of self-supervised learning, Vision-Language Pre-training has emerged as a prominent topic. CLIP~\cite{CLIP} initially demonstrated that equipped with large-scale image-text pairs, a contrastive learning framework can achieve a performance comparable to fully supervised baselines. Subsequently, ALIGN~\cite{DBLP:conf/icml/JiaYXCPPLSLD21} scales the training dataset to billions and gains a better vision-language representation. Another line of work~\cite{DBLP:conf/nips/Gan0LZ0020, DBLP:journals/corr/abs-1909-11740} leverage pre-trained object detection models like Faster-RCNN~\cite{DBLP:conf/nips/RenHGS15} to extract image regional features offline for training multi-modal transformers. These models show impressive performance on various vision-language tasks, as they are capable of processing both single-modal and multi-modal inputs. In this paper, we make the first attempt to leverage the powerful CLIP for the UCDR task.

\subsection{Prompt Tuning}

Prompt Tuning~\cite{DBLP:journals/csur/LiuYFJHN23}, as a new paradigm, initially surfaces in NLP for adapting pre-trained language models (PLMs)~\cite{DBLP:conf/naacl/DevlinCLT19,DBLP:conf/nips/BrownMRSKDNSSAA20} to downstream tasks. These prompts can either be handcrafted for specific tasks~\cite{DBLP:conf/emnlp/ShinRLWS20} or learned automatically via gradients which are termed as ``Prompt Learning''~\cite{DBLP:conf/emnlp/LesterAC21}. Driven by the success of prompt paradigm in NLP, various studies like CoOP~\cite{coco} and CoCoOp~\cite{cocoop} apply text prompting in multi-modal scenarios, while VPT~\cite{vpt} and VP~\cite{vp123} introduce this paradigm into vision transformer. Despite the pioneering successes of VP and VPT in visual models, we find that they are static after training which is not flexible to handle the generalized scenarios. In contrast, our method proposes a Prompt-to-Simulate method to well address UCDR task.

\section{Method}
\begin{figure*}
    \centering
    \includegraphics[width=0.9\linewidth]{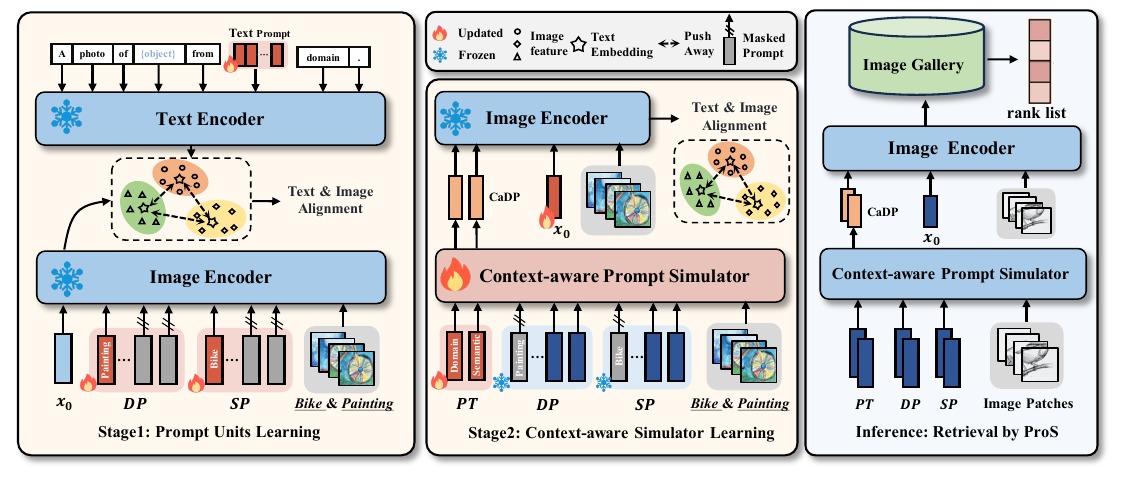}
    \vspace{-1em}
    \caption{\textbf{Overview of our proposed ProS}. In Prompt Units Learning Stage, we capture knowledge from source data into domain prompts units $DP$ and semantic prompts units $SP$ by masking irrelevance prompts. In the Context-aware Prompt Simulation Stage, we train a Context-aware Prompt Simulator (CaPS) with a mask operation to dynamically convey prompt templates $PT$ to two Content-aware Dynamic Prompts (CaDP) to simulate unknown domains and categories. In the retrieval phase, we employ CaPS to produce CaDP which impacts the CLIP image encoder to convert unseen samples into suitable embeddings for retrieval. The gray parts indicate masked prompts.}
    \label{fig:training}
    \vspace{-1em}
\end{figure*}

Our method makes the first attempt to apply CLIP for Universal Cross-Domain Retrieval (UCDR) with a novel prompt learning method named ProS, as shown in \cref{fig:training}. We begin with a brief overview of some important preliminaries about UCDR, CLIP, and prompt tuning paradigm (\cref{sec:Preliminary}). Then, a detailed description of our proposed Prompting-to-Simulate (Pros) is provided (\cref{sec:Prompt-to-Simulate}), including two training stages: Prompt Unit Learning (PUL) and Context-aware Simulator Learning (CSL). Finally, we will introduce how to perform retrieval using our proposed method (\cref{sec:inference time}).

\subsection{Preliminary}

\label{sec:Preliminary}

\textbf{Problem Formulation.}
Universal Cross-Domain Retrieval (UCDR)~\cite{UCDR} aims to accurately retrieve images with queries from unseen domains and classes. From a general perspective, we assume that the training dataset is composed of $K$-different ($K\ge2$) source domains (Real, Sketch, Quickdraw, \etc) which can be formulated as $\mathcal{D}_{train }=\bigcup_{d \in\left\{1, \ldots, K\right\}}\left\{\boldsymbol{x}_i^d , y^{tr}_i \right\}_{i=1}^{N_{d}}$. The $d^{th}$ domain has $N_d$ images, where each image $\boldsymbol{x}_i^d$ belongs to  class $y^{tr}_i\in \mathcal{C}_{train }$. In the testing phase, the gallery set and query set are needed to enable image retrieval. Specifically, the query set $\mathcal{D}_{query}=\left\{\boldsymbol{x}_i^q , y^{te}_i\right\}_{i=1}^{N_{q}}$ consists of $N_q$ samples belonging to unseen classes $y_i^{\text{te}} \in \mathcal{C}_{\text{test}} $ from the query domain $q$, while the gallery set consists of real images. To simulate a more realistic retrieval scenario, the gallery set has two settings: 1) all samples belong to the unseen class, termed \textit{Unseen Gallery}; (2) samples belong to both seen class and unseen class, termed \textit{Mixed Gallery}.

The UCDR protocol can be separated into two sub-tasks, namely: (a) $\mathrm{U}^{c} \mathrm{CDR}$, where the query domain $q \in \{1,...,K\}$ but the classes are held out, \ie $\mathcal{C_{\text{train}}} \cap \mathcal{C_{\text{test}}}  \equiv \phi$. (b) $\mathrm{U}^{d} \mathrm{CDR}$, where the query domain $q \notin \{1,...,K\}$ but the classes are seen, \ie $\mathcal{C_{\text{train}}} \equiv  \mathcal{C_{\text{test}}}$. The goal of UCDR is to learn a mapping function that can place query and gallery images into a latent domain-independent subspace $\Phi$ , where images of the same label but different domains are clustered together. The most challenging part is images from unseen domains and unknown classes also need to be mapped correctly in $\Phi$.

\noindent \textbf{CLIP.} Contrastive Language-Image Pre-training (CLIP) \cite{CLIP} is a multi-modality model pre-trained from 400 million image-text pairs. It aligns texts and images by an image encoder $f(\cdot)$ and a text encoder $g(\cdot)$. 
CLIP classifies images in a zero-shot way, based on the similarity between image features $f(\boldsymbol{x})$ and text features. The text features of various class-wise caption $\left\{\boldsymbol{t}_i\right\}_{i=1}^{|C_{train}|}$ can be formulated as $\left\{g(\boldsymbol{t}_i)\right\}_{i=1}^{|C_{train}|}$, which are produced manually by filling class names into a text template. For example,  given a text template  ``a photo of a [CLASS].", then exchange [CLASS] with class names to construct class-wise caption, \eg, [CLASS] $\rightarrow$ cat / dog / bike.  Formally speaking, given an image $\boldsymbol{x}$ and class caption $\left\{\boldsymbol{t}_i\right\}_{i=1}^{|C_{train}|}$ , CLIP output a prediction by:
\begin{equation}
    \vspace{-1em}
  \hat{y}_{clip}=\underset{i}{\arg \max} (f(\boldsymbol{x})\otimes g(\boldsymbol{t}_i)),
\end{equation}
where $\otimes $ is cosine similarity. Benefiting from large-scale pre-training and alignment of images and text, CLIP is robust for various visual appearance changes~\cite{Pretrained_Improve_Robustness}, which is suitable for UCDR. However, fine-tuning CLIP will distort pre-trained features and lead to poor out-of-distribution performance~\cite{Fine-Tuning_Distort_Features}. Therefore, how to effectively leverage this powerful foundation model for UCDR becomes an important problem.

\noindent \textbf{Prompt Tuning.} Compared with full fine-tuning, prompt tuning prevents pre-trained feature distortion by introducing task-specific trainable parameters termed prompts into the input while keeping the pre-trained model frozen. Specifically, \cite{coco} incorporate learnable text prompts $\mathbf{P_{t}}=\left\{p_t^i \in \mathbb{R}^{\ell
} \right\}_{i=1}^{N_{t}}$ into the text template as the input of CLIP text encoder, where $N_t$ and $ \mathbb{R}^{\ell}$ indicate text prompts length and dimensions, respectively. Then the text template can be represented as  ``$\mathbf{P_{t}}$ a photo of a [CLASS]". Furthermore,  \cite{vpt} proposed visual prompts for ViT which can be formulated as $\mathbf{P_{v}}=\left\{p_v^i \in \mathbb{R}^{\ell
} \right\}_{i=1}^{N_{v}}$, where $N_{v}$ represents the number of visual prompts and $p_v$ is a learnable parameter with $\ell$ dimension. Then the visual prompts are inserted into the first Transformer layer, named VPT-Shallow in \cite{vpt} :
\begin{equation}
    \label{prompted encoder}
    \begin{aligned}
    {\left[\mathbf{x}_{1}, \mathbf{Z}_{1}, \mathbf{E}_{1}\right] } & =L_{1}\left(\left[\mathbf{x}_{0}, \mathbf{P_{v}}, \mathbf{E}_{0}\right]\right), \\
    {\left[\mathbf{x}_{i}, \mathbf{Z}_{i}, \mathbf{E}_{i}\right] } & =L_{i}\left(\left[\mathbf{x}_{i-1}, \mathbf{Z}_{i-1}, \mathbf{E}_{i-1}\right]\right) \quad i=2,3, \ldots, N_L, \\
    \mathbf{Output} & =\operatorname{Head}\left(\mathbf{x}_{N_L}\right),
    \end{aligned}
\end{equation}
where $\mathbf{Z}_i \in \mathbb{R}^{N_{v} \times \ell}$ represents the prompt embeddings computed by the $i^{th}$ Transformer layer in the CLIP image encoder, $\mathbf{E}_{0}$ means the image patch embeddings extracted from a CNN feature map~\cite{ViT}. In addition, $\mathbf{x_0}$ is \textit{CLS} token, which produces $\mathbf{y}$ and serves as the final output of image representation.

\subsection{Prompt-to-Simulate}
\label{sec:Prompt-to-Simulate}

\noindent \textbf{Prompt Units Learning (PUL).} To capture domain-specific and class-specific knowledge in PUL, we introduce two types of prompt units: domain prompt units $DP$ and semantic prompt units $SP$, as shown in~\cref{fig:training} (left). Specifically, we employ $K$ learnable vectors as domain prompt units to represent $K$ source domains in training datasets:
\begin{equation}
    \label{specific domain prompt}
    DP = \left\{dp_i \in \mathbb{R}^{\ell} \right\}_{i=1}^K.
\end{equation}
Similarly, we introduce a set of learnable semantic prompts for $\left|\mathcal{C_{\text{train}}}\right|$ seen classes which can be formulated as:
\begin{equation}
    \label{specific class prompt}
    SP = \left\{sp_i \in \mathbb{R}^{\ell} \right\}^{\left|\mathcal{C_{\text{train}}}\right|}_{i=1}.
\end{equation}
To ensure that each prompt only learns knowledge related to a specific domain or class, we iteratively update single $dp$ and $sp$ \wrt an input image. To achieve this, we mask the \textbf{i}rrelevant \textbf{d}omain and \textbf{s}emantic prompt units by $\mathbf{M_{id}}=\left\{\mathbf{m_{id}}^i \in [0,1] \right\}^{K}_{i=1}$ and $\mathbf{M_{is}} = \left\{\mathbf{m_{is}}^i \in  [0,1] \right\}^{\left|\mathcal{C_{\text{train}}}\right|}_{i=1}$. Assume that, the input image is from domain $\mathbf{d}$ and class $\mathbf{c}$, than $\mathbf{m_{id}^d}$ and $\mathbf{m_{is}^c}$ are $1$ and other entries are $0$. After mask operation, two masked prompt units are then concatenated with other input tokens to formulate a prompted input: 
\begin{equation}
    \label{prompted input}
    \mathbf{x_{p}}= [\mathbf{x_0},  \mathbf{M_{id}} \odot DP,  \mathbf{M_{is}} \odot SP, \mathbf{E_0}],
\end{equation}
where $\odot$ discards prompts with corresponding mask of $0$ preserves others of $1$. Correspondingly, the image feature is obtained in the same way as \cref{prompted encoder}. 
Since the text feature can be considered as the class prototype, we introduce the learnable text prompts $\mathbf{P_{t}}$ as we discussed in \cref{sec:Preliminary} to achieve more flexible training. Additionally, we design a new text template: ``a photo of $[CLASS]^i$ from $\mathbf{P_{t}}$ domain.", to better suit the UCDR task. Given the prompted input $\mathbf{x_{p}}$ and the class caption $\left\{\boldsymbol{t}_i\right\}_{i=1}^{|C_{train}|}$ generated by above text template, the training loss can be formulated as:
\begin{equation}
    \label{SPL loss}
  \mathcal{L} = -\sum^{\left|\mathcal{C_{\text{train}}}\right|}_{i=1}  y_i \log{( f(\mathbf{\mathbf{x_p}})\otimes g\left(\boldsymbol{t}_i \right))},
\end{equation}
where $\otimes$ is cosine similarity and $y$ indicates the class label. The above procedure results in a mask-and-align objective. Our masked input ensures prompts to capture specific source knowledge from domains and categories. Then, the image features and text features are aligned among different domains in a union semantic space.

\noindent \textbf{Context-aware Simulator Learning (CSL).} In CSL, our goal is to train a module termed Content-aware Prompt Simulator (CaPS) to generate two Content-aware Dynamic Prompts (CaDP) similar to Prompt Units which can well impact CLIP to extract generalized features of unseen samples. Towards this goal, we concatenate two sets of Prompt Units with image patches as CaPS's input, which can be formulated as $[PT_d, PT_s, DP, SP, \mathbf{E_0}]$, where $PT_d \in \mathbb{R}^{\ell},\;PT_s \in \mathbb{R}^{\ell}$ are domain and semantic prompt template for CaDP generation and $\mathbf{E_0}$ are unseen sample's embedding. 

Inspired by the idea of meta-learning, we train CaPS under a simulated test scenario by a relevant mask operation. Specifically, we mask \textbf{r}elevant \textbf{d}omain and \textbf{s}emantic prompt units \wrt input samples by $\mathbf{M_{rd}}=\left\{\mathbf{m_{rd}}^i \in [0,1] \right\}^{K}_{i=1}$ and $\mathbf{M_{rs}} = \left\{\mathbf{m_{rs}}^i \in  [0,1] \right\}^{\left|\mathcal{C_{\text{train}}}\right|}_{i=1}$, where  $\mathbf{m_{rd}^d} = 0$, $\mathbf{m_{rs}^c} = 0$ and others are $1$ if the input sample is from domain $\mathbf{d}$ and class $\mathbf{c}$. Then we can hide the prompt units corresponding to input samples by multiplying them with $\mathbf{M_{rd}}$ and $\mathbf{M_{rs}}$ mask matrices. The whole process can be formulated as:
\begin{equation}
    \label{encoder forwarded}
    \begin{aligned}
    {\left[\mathcal{P}_d, \mathcal{P}_s\right]}&= \mathcal{M}([PT_d, PT_s, \mathbf{M_{rd}} \odot DP, \mathbf{M_{rs}} \odot SP, \mathbf{E_0}]),\\
    \mathbf{x_{p}} &= [\mathbf{x_0},  \mathcal{P}_d, \mathcal{P}_s, \mathbf{E_0}],
    \end{aligned}
\end{equation}
Similarly, we use the same objective as \cref{SPL loss} to train the newly introduced $PT_d, PT_s$ and $\mathcal{M}$ while the remaining part would be fixed in this stage.

\subsection{Retrieval by ProS}
\label{sec:inference time}

After two-stage training, now we can generate CaDP by CaPS, and extract image features by CLIP with the help of CaPS.\cref{fig:training} (right) shows how to employ our proposed ProS for retrieval. Since we only need image features for retrieval, CLIP text encoder $g$ is discarded. We first use CaPS $\mathcal{M}$  to generate two CaDP, and then use CLIP Image Encoder $f$ to get well-generalized image features $\mathbf{y}$:
\begin{equation}
    \label{infernece}
    \begin{aligned}
    {\left[\mathcal{P}_d, \mathcal{P}_s\right]}&= \mathcal{M}([PT_d, PT_s, DP, SP, \mathbf{E_0}]),\\
   \mathbf{x} & =f\left(\left[\mathbf{x_0}, \mathcal{P}_d, \mathcal{P}_s, \mathbf{E}_{0}\right]\right), \\
    \mathbf{y} & =\operatorname{Head}\left(\mathbf{x}\right).
    \end{aligned}
\end{equation}
To perform retrieval, all images in $\mathcal{C}_{\text{test}}$ are indexed by \cref{infernece} to formulate the gallery. Then, given a query, we do the same thing to obtain the query feature and produce a rank list by sorting similarities between the query and all features in the gallery. In the next section, we will give a comprehensive study of the retrieval results of our proposed ProS.

\begin{table*}[]
\centering
\scalebox{0.85}{
\begin{tabular}{cccccccc}
\toprule
\multirow{3}{*}{\begin{tabular}[c]{@{}c@{}}Query\\ Domain\end{tabular}} & \multirow{3}{*}{Method} & \multicolumn{4}{c}{UCDR}                                                                 & \multicolumn{2}{c}{$\mathrm{U^d} \mathrm{CDR}$}          \\ \cmidrule(r){3-6} \cmidrule(r){7-8}
                                                                        &                         & \multicolumn{2}{c}{Unseen Gallery} & \multicolumn{2}{c}{Mixed Gallery} & \multirow{2}{*}{mAP@200} & \multirow{2}{*}{Prec@200} \\ \cmidrule(r){3-4}  \cmidrule(r){5-6}
                                                                        &                         & mAP@200             & Prec@200           & mAP@200               & Prec@200              &                          &                           \\  \midrule
\multirow{6}{*}{Sketch}                                                 & SnMpNet~\cite{UCDR} & 0.3007              & 0.2432             & 0.2624                & 0.2134                & 0.3529                   & 0.1657                    \\
                                                                        & SASA~\cite{SASA}   & 0.5262              & 0.4468             & 0.4732                & 0.4025                & 0.5733                   & 0.5290                     \\
                                                                        & CLIP-Full               & 0.5367              & 0.4666             & 0.4788                & 0.4136                & 0.6128                   & 0.3806                    \\
                                                                        & CoOp~\cite{coco}     & 0.5512              & 0.4947             & 0.4995                & 0.4479                & 0.6374                   & 0.4245                    \\
                                                                        & VPT~\cite{vpt}      & 0.6216              & 0.5676             & 0.5609                & 0.5130                 & 0.6769                   & 0.4405                    \\
                                                                        & \textbf{ProS (Ours)}     & \textbf{0.6457}     & \textbf{0.6001}    & \textbf{0.5843}       & \textbf{0.5463}       & \textbf{0.7385}          & \textbf{0.4911}           \\ \hline
\multirow{6}{*}{Quickdraw}                                              & SnMpNet~\cite{UCDR} & 0.1736              & 0.1284             & 0.1512                & 0.1111                & 0.1077                   & 0.0509                    \\
                                                                        & SASA~\cite{SASA}   & 0.2564              & 0.1970              & 0.2116                & 0.1651                & 0.1805                   & 0.1549                    \\
                                                                        & CLIP-Full               & 0.2011              & 0.1522             & 0.1622                & 0.1196                & 0.1820                    & 0.0723                    \\
                                                                        & CoOp~\cite{coco}     & 0.1484              & 0.1237             & 0.1183                & 0.0961                & 0.1834                   & 0.084                     \\
                                                                        & VPT~\cite{vpt}      & 0.2467              & 0.2092             & 0.1953                & 0.1688                & 0.2367                   & 0.0982                    \\
                                                                        & \textbf{ProS (Ours)}     & \textbf{0.2842}     & \textbf{0.2544}    & \textbf{0.2318}       & \textbf{0.2127}       & \textbf{0.2889}          & \textbf{0.1186}           \\ \hline
\multirow{6}{*}{Painting}                                               & SnMpNet~\cite{UCDR} & 0.4031              & 0.3332             & 0.3635                & 0.3019                & 0.4808                   & 0.4424                    \\
                                                                        & SASA~\cite{SASA}   & 0.5898              & 0.5188             & 0.5463                & 0.4804                & 0.5596                   & 0.5178                    \\
                                                                        & CLIP-Full               & 0.6558              & 0.5926             & 0.6083                & 0.5478                & 0.6189                   & 0.3688                    \\
                                                                        & CoOp~\cite{coco}     & 0.6886              & 0.6207             & 0.6509                & 0.5884                & 0.6625                   & 0.4128                    \\
                                                                        & VPT~\cite{vpt}      & 0.7138              & 0.6503             & 0.6752                & 0.6153                & 0.6618                   & 0.4105                    \\
                                                                        & \textbf{ProS (Ours)}     & \textbf{0.7516}     & \textbf{0.6955}    & \textbf{0.7120}        & \textbf{0.6612}       & \textbf{0.7227}          & \textbf{0.4615}           \\ \hline
\multirow{6}{*}{Infograph}                                              & SnMpNet~\cite{UCDR} & 0.2079              & 0.1717             & 0.1800                  & 0.1496                & 0.1957                   & 0.1764                    \\
                                                                        & SASA~\cite{SASA}   & 0.2823              & 0.2425             & 0.2491                & 0.2113                & 0.2340                    & 0.2093                    \\
                                                                        & CLIP-Full               & 0.5332              & 0.4893             & 0.4718                & 0.4309                & 0.5311                   & 0.3330                     \\
                                                                        & CoOp~\cite{coco}     & 0.5285              & 0.4807             & 0.4820                 & 0.4390                 & 0.5530                    & 0.3546                    \\
                                                                        & VPT~\cite{vpt}      & 0.5434              & 0.4957             & 0.4870                 & 0.4468                & 0.5690                    & 0.3566                    \\
                                                                        & \textbf{ProS (Ours)}     & \textbf{0.5798}     & \textbf{0.5442}    & \textbf{0.5219}       & \textbf{0.4956}       & \textbf{0.6056}          & \textbf{0.3962}           \\ \hline
\multirow{6}{*}{Clipart}                                                & SnMpNet~\cite{UCDR} & 0.4198              & 0.3323             & 0.3765                & 0.2959                & 0.552                    & 0.5074                    \\
                                                                        & SASA~\cite{SASA}   & 0.4397              & 0.3670              & 0.3940                 & 0.3295                & 0.684                    & 0.6361                    \\
                                                                        & CLIP-Full               & 0.6880               & 0.6200               & 0.6423                & 0.5755                & 0.6922                   & 0.4174                    \\
                                                                        & CoOp~\cite{coco}     & 0.7025              & 0.6414             & 0.6648                & 0.6068                & 0.7495                   & 0.4776                    \\
                                                                        & VPT~\cite{vpt}      & 0.7344              & 0.6785             & 0.6942                & 0.6409                & 0.7536                   & 0.4770                     \\
                                                                        & \textbf{ProS (Ours)}     & \textbf{0.7648}     & \textbf{0.7186}    & \textbf{0.7228}       & \textbf{0.6815}       & \textbf{0.8105}          & \textbf{0.5298}           \\ \hline
\multirow{6}{*}{Average}                                                & SnMpNet~\cite{UCDR} & 0.3010             & 0.2418            & 0.2667               & 0.2144               & 0.3378                  & 0.2686                   \\
                                                                        & SASA~\cite{SASA}   & 0.4189             & 0.3544            & 0.3748               & 0.3178               & 0.4463                  & 0.4094                   \\
                                                                        & CLIP-Full               & 0.5229             & 0.4641            & 0.4727               & 0.4175               & 0.5274                   & 0.3144                   \\
                                                                        & CoOp~\cite{coco}     & 0.5238             & 0.4722            & 0.4831                & 0.4356               & 0.5572                  & 0.3507                    \\
                                                                        & VPT~\cite{vpt}      & 0.5720             & 0.5203            & 0.5225               & 0.4770               & 0.5796                   & 0.3566                   \\
                                                                        & \textbf{ProS (Ours)}     & \textbf{0.6052}    & \textbf{0.5626}   & \textbf{0.5546}      & \textbf{0.5195}      & \textbf{0.6332}         & \textbf{0.3994}          \\ \bottomrule
\multicolumn{1}{l}{}                                                    & \multicolumn{1}{l}{}    &                     &                    &                       &                       & \multicolumn{1}{l}{}     & \multicolumn{1}{l}{}     
\end{tabular}
}
\vspace{-1.5em}
\caption{UCDR and $\mathrm{U^d CDR}$ evaluation results on DomainNet. UCDR has two different gallery settings, \ie the gallery set consists of (1) only unseen class images (Unseen Gallery) or (2) both seen and unseen images from Real domain (Mixed Gallery).} 
\vspace{-1.5em}
\label{tab1}
\end{table*}

\section{Experiment}

\subsection{Experimental setting}

\textbf{Datasets.} 
Following the existing works~\cite{UCDR,SASA}, we evaluate the effectiveness of our ProS on three popular datasets, including DomainNet~\cite{domainet}, Sketchy~\cite{sketchy, sketchy_tuberlin}, and TU-Berlin~\cite{tuberlin,sketchy_tuberlin}) under three different cross-domain retrieval settings, \ie, UCDR, $\mathrm{U}^{c} \mathrm{CDR}$ and $\mathrm{U}^{d} \mathrm{CDR}$ as we introduced in~\cref{sec:Preliminary}.

\textbf{DomainNet}~\cite{domainet} contains 596,006 images collected in 6 domains (\textit{Real, Sketch, Quickdraw, Infograph, Clipart, Painting}) from 345 categories and split into 245, 55, 45 categories for training, validation, and testing respectively. To satisfy the unseen domain requirement in both $\mathrm{U^d CDR}$ and $\mathrm{UCDR}$ protocol, \textit{leave-one-out} evaluation protocol is applied, where we iteratively select one domain as unseen query set and use other domains for training. Additionally, for $\mathrm{U^d CDR}$ evaluation, we select 45 categories from training set and choose $25\%$ samples from unseen domains as queries (except for Quickdraw, which is $10\%$ due to its large size). The gallery set is constructed with \textit{Real} images from \textit{unseen} categories or mixed with \textit{seen} categories, termed \textit{Unseen Gallery} and \textit{Mixed Gallery} respectively.

\textbf{Sketchy}~\cite{sketchy, sketchy_tuberlin} contains 75,471 Sketches and 73,002 images from 125 categories where the train, validation, and test splits contain 93, 11, and 21 categories~\cite{SASA,UCDR}, respectively. \textbf{TU-Berlin}~\cite{tuberlin,sketchy_tuberlin} has 20,000 Sketches and 204,489 images which are separated into 200 categories for training, 20 categories for validation, and 30 categories for testing~\cite{SASA,UCDR}. Both Sketchy and TU-Berlin are used for $\mathrm{U^c CDR}$.

\noindent \textbf{Evaluation Metrics.}
For a fair comparison, we utilize the same evaluation metrics following \cite{UCDR,SASA}. For Sketchy and DomainNet, we evaluate precision and mean Average Precision \wrt top-200 candidates (Prec@200 and mAP@200). For TU-Berlin, we use Prec@100 and mAP@all as evaluation metrics instead.

\noindent \textbf{Implementation Details.}
In all our experiments, we use a fixed image encoder and text encoder initialized with pre-trained CLIP (ViT-B/32). For Context-aware Prompt Simulator $\mathcal{M}$, we adopt a two-layer ViT with random initialization. Following ~\cite{coco}, the length of text prompt $\mathbf{P_{t}}$ is set to 16 with 512 dimensions. In addition,  the dimensions of other prompts including prompt units and Content-aware Dynamic Prompts are set to 768.  The training epochs are set to 10 with early stopping of 2 epochs based on the evaluation performance for all the datasets with a batch size of 50. We use Adam optimizer with a learning rate of 1e-3 and adopt a cosine decay strategy for the learning rate.

\subsection{Main Results}
\noindent \textbf{Baselines.} We compare our ProS with two groups of methods, including two existing UCDR methods, \ie, SnMpNet~\cite{UCDR} which adopt a ResNet as the backbone and SASA~\cite{SASA}, which implement with a ViT. For fair evaluation, we further build three CLIP-based baselines, including CLIP-full and two prompt tuning methods, \ie, CoOp~\cite{coco} and VPT~\cite{vpt} where CLIP-full indicates fine-tuning the full CLIP.

\noindent \textbf{Results on UCDR and $\mathbf{U^d CDR}$.} We first evaluate our ProS methods on DomainNet under UCDR and $\mathrm{U^d} \mathrm{CDR}$ settings. The UCDR results in \cref{tab1} show that our ProS method outperforms all baselines across all domains, demonstrating the superiority of ProS in terms of generalization on unseen domains and categories. Furthermore, we have highlighted the following conclusions: \textbf{First} (CLIP \vs ViT): all CLIP-based methods outperform ViT-based method which proves our assumption that general knowledge gained in large-scale pre-training can enhance the model's ability to handle the UCDR task. Specifically, CLIP-Full achieves 9.79\% mAP@200 improvement than SASA. \textbf{Second} (Prompt Tuning \vs Fine-Tuning): VPT and CoOp both outperform CLIP-Full, demonstrating the effectiveness of prompt tuning to adapt CLIP. \textbf{Third} (ProS \vs Prompt Tuning): compared with VPT, our ProS further improves the performance by 3.21\% mAP@200 on average, indicating that ProS effectively applies prompt tuning for UCDR.

From reported $\mathbf{U^d CDR}$ results in \cref{tab1}, we can draw the conclusion that our ProS consistently outperforms all competitors in terms of cross-domain alignment and generalization. Specifically, comparing with average results, our method surpasses VPT by $5.36\%$ in mAP@200.

\begin{table}[]
\centering
\scalebox{0.8}{
\begin{tabular}{ccccc}
 \toprule
\multirow{2}{*}{Method} & \multicolumn{2}{c}{Sketchy}                              & \multicolumn{2}{c}{TU-Berlin}                            \\ \cmidrule(r){2-3}  \cmidrule(r){4-5}
                       & mAP@200                            & Prec@200                            & mAP@all                             & Prec@100                            \\ \midrule
SnMpNet~\cite{UCDR}                 & 0.5781                              & 0.5155                              & 0.3568                              & 0.5226                              \\
SASA~\cite{SASA}                    & 0.6910                               & 0.6090                               & 0.4715                              & 0.6682                              \\
CLIP-Full               & 0.6553                              & 0.6145                              & 0.6076                              & 0.7158                              \\
CoOp~\cite{coco}                    & \multicolumn{1}{c}{0.5074}          & \multicolumn{1}{c}{0.4659}          & \multicolumn{1}{c}{0.5585}          & \multicolumn{1}{c}{0.6759}          \\
VPT~\cite{vpt}                     & \multicolumn{1}{c}{0.6588}          & \multicolumn{1}{c}{0.6105}          & \multicolumn{1}{c}{0.5574}          & \multicolumn{1}{c}{0.6815}          \\
\textbf{ProS (Ours)}           & \multicolumn{1}{c}{\textbf{0.6991}} & \multicolumn{1}{c}{\textbf{0.6545}} & \multicolumn{1}{c}{\textbf{0.6675}} & \multicolumn{1}{c}{\textbf{0.7442}} \\ \bottomrule
\end{tabular}
}
\caption{$\mathrm{U^c CDR}$ evaluation results on Sketchy and TU-Berlin. Consistent with \cite{UCDR, SASA}, we use Prec@200 and mAP@200 in Sketchy, and Prec@100 and mAP@all in TU-Berlin as evaluation metrics.}
\vspace{-1em}
\label{tab:UcCDR}
\end{table}

\noindent \textbf{Results on $\mathbf{U^c CDR}$.}
In \cref{tab:UcCDR}, we explore the ability of ProS in handling semantic shift under $\mathrm{U^c CDR}$ setting on Sketchy and TU-Berlin. We observe that: \textbf{First}, our ProS shows consistent improvement compared to CLIP-Full and VPT, revealing that our ProS can enhance the ability of CLIP to manage semantic shifts and generalize to unseen categories. \textbf{Second,} compared to SASA, our ProS has a larger improvement on TU-Berlin than that on Sketchy($19.6\%$ \vs $0.81\%$ in mAP@200). We could see CLIP, CoOp, and VPT,  all of which use CLIP as the backbone, exhibit relatively low mAP on Sketchy, potentially caused by poor performance of the visual backbone when handling this dataset. However, our method still surpasses all competitors and achieves the best performance.

\subsection{Ablation Study}
In this section, we conduct extensive ablation experiments to reveal the contributions of each component. For all ablations, we select Infograph as unseen query set from DomainNet and evaluate ProS under UCDR setting with two gallery configurations.

\noindent \textbf{Impact of Each Components.}
To validate the effectiveness of each component in our ProS, we individually remove each component and experiment with the following five settings: 1) ``w/o $SP$" (remove Semantic Prompt Units). 2) ``w/o" $DP$ (remove Domain Prompt Units). 3) ``w/o Mask" (remove mask operation in training). 4) ``w/o $\mathcal{M}$" (remove Content-aware Prompt Simulator). 5) ``w/o \textit{CLS}" (remove CLS token). The experiment results are shown in \cref{tab:Component}. We observe that: \textbf{ First,} removing either $SP$ or $DP$ led to a performance degradation of -2.65\% and -2.56\% in mAP@200 under mixed gallery settings, indicating their essential roles in Content-aware Prompt Simulation stage. \textbf{Second,} without the mask operation, mAP@200 has decreased by 1.27\% which proves the effectiveness of mask operation. \textbf{Third,} when we remove CaPS $\mathcal{M}$ from our model, it leads to worse performance than full ProS, highlighting the significant benefits of $\mathcal{M}$ for the UCDR task. \textbf{Forth,} training $CLS$ token is absolutely necessary as its omission caused a performance reduction of 6.18

\begin{table}[]
\centering
\scalebox{0.85}{
\begin{tabular}{lcccc}
\toprule
\multirow{2}{*}{Method} & \multicolumn{2}{c}{Unseen Gallery}                   & \multicolumn{2}{c}{Mixed Gallery}              \\ \cmidrule(r){2-3}  \cmidrule(r){4-5}
                        & \multicolumn{1}{c}{mAP@200} & \multicolumn{1}{c}{Prec200} & \multicolumn{1}{c}{mAP200} & \multicolumn{1}{c}{Prec200} \\ \midrule
\textbf{ProS}     & \textbf{0.5798}             & \textbf{0.5442}              & \textbf{0.5219}             & \textbf{0.4956}              \\
w/o $SP$                  & 0.5529                      & 0.5142                       & 0.4954                      & 0.4641                       \\
w/o $DP$                  & 0.5566                      & 0.5190                        & 0.4963                      & 0.4680                      \\
w/o Mask                & 0.5692                      & 0.5324                       & 0.5092                      & 0.4809                       \\
w/o $\mathcal{M}$       & 0.5573                      & 0.5128                        & 0.5053                      & 0.4685                      \\ 
w/o $CLS$                 & 0.5241                      & 0.4846                       & 0.4601                      & 0.4260                        \\ \bottomrule

\end{tabular}
}
\caption{Different components evaluated on DomainNet under UCDR protocol with Infograph as the unseen domain for queries. }
\label{tab:Component}
\end{table}

\noindent \textbf{Impact of Transformer Layer in $\mathcal{M}$.}
We further investigate the impact of Transformer layers in Context-aware Prompt Simulator $\mathcal{M}$. From the~\cref{tab:Transformer Layer}, it can be found that: 2 layer ViT yields the best performance. Although employing only 1 layer can slightly reduce computational costs, it results in a worse performance. Hence, for an optimal trade-off between accuracy and efficiency,  a 2-layer ViT is chosen to implement $\mathcal{M}$.

\begin{table}[]
\centering
\scalebox{0.85}{
\begin{tabular}{ccccc}
\toprule
\multirow{2}{*}{$\mathcal{M}$'s Layer}  & \multicolumn{2}{c}{Unseen Gallery} & \multicolumn{2}{c}{Mixed Gallery} \\ \cmidrule(r){2-3} \cmidrule(r){4-5}
                                                         & mAP@200         & Prec@200         & mAP@200         & Prec@200        \\ \midrule
1                                               & 0.5752          & 0.5379           & 0.5177          & 0.4900          \\
2                                              & 0.5798          & 0.5442           & 0.5219          & 0.4956          \\
3                                            & 0.5790          & 0.5429           & 0.5210          & 0.4942          \\
6                                           & 0.5743          & 0.5388           & 0.5160          & 0.4893          \\
12                                        & 0.5598          & 0.5228           & 0.5036          & 0.4742          \\ 
\bottomrule
\end{tabular}
}
\caption{Ablation study on the effectiveness of different transformer layers in Context-aware Prompt Simulator $\mathcal{M}$.}
\label{tab:Transformer Layer}
\end{table}
\vspace{-2em}
\begin{figure}
    \centering
    \includegraphics[width=1\linewidth]{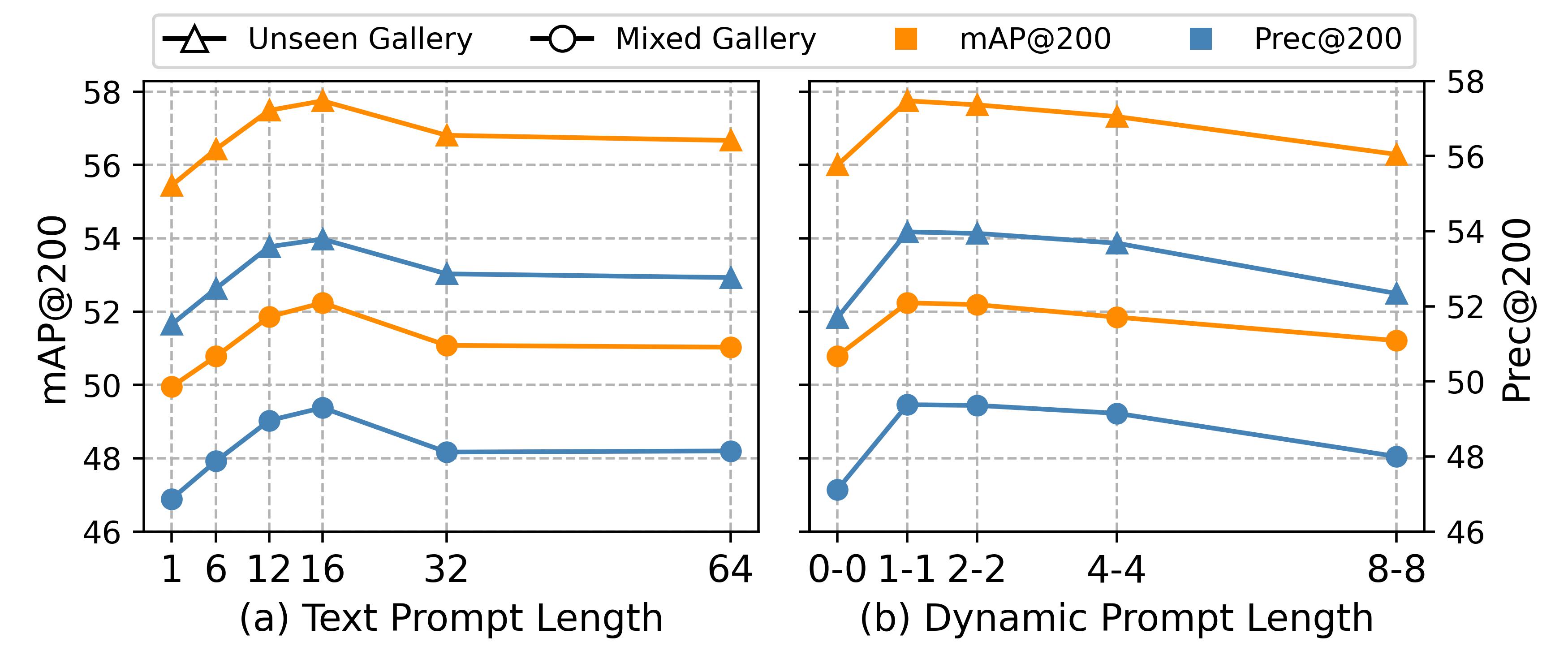}

    \caption{Evaluation results of two prompts length. (a) investigate the impact of text prompt length. (b) analyze Content-aware Dynamic Prompt length generated by CaPS $\mathcal{M}$, where 0-0 represents VPT and 1-1 means one CaDP for domain and one for semantic.}
     \vspace{-2em}
    \label{fig:prompt length}
\end{figure}

\begin{figure*}
    \centering
    \includegraphics[width=0.9\linewidth]{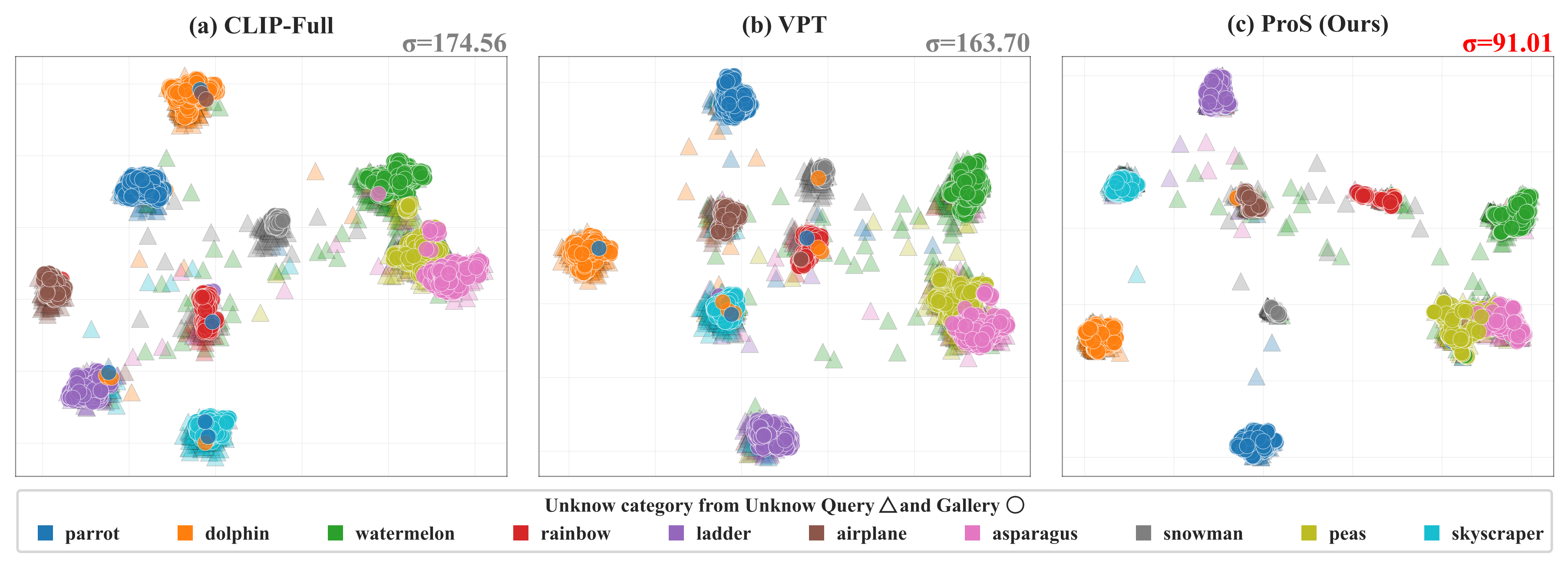}
    \vspace{-1em}
    \caption{Visualization of image features from 10 randomly selected unseen classes of \textit{Real} Query and unseen \textit{Infograph} Gallery set. Different colors represent different categories while $\bigcirc$ and $\bigtriangleup $ represent samples from \textit{real} and \textit{Infograph} domains, respectively. We further evaluate performance by metric from \cite{LBHash}, \ie, $\sigma=\frac{\max\mathcal{D_{\mathit{intra}}}}{\min\mathcal{D_{\mathit{inter}}}}$ (lower is better).}
    \label{fig:visualization}
    \vspace{-1.5em}
\end{figure*}

\begin{figure}
    \centering
    \includegraphics[width=1\linewidth]{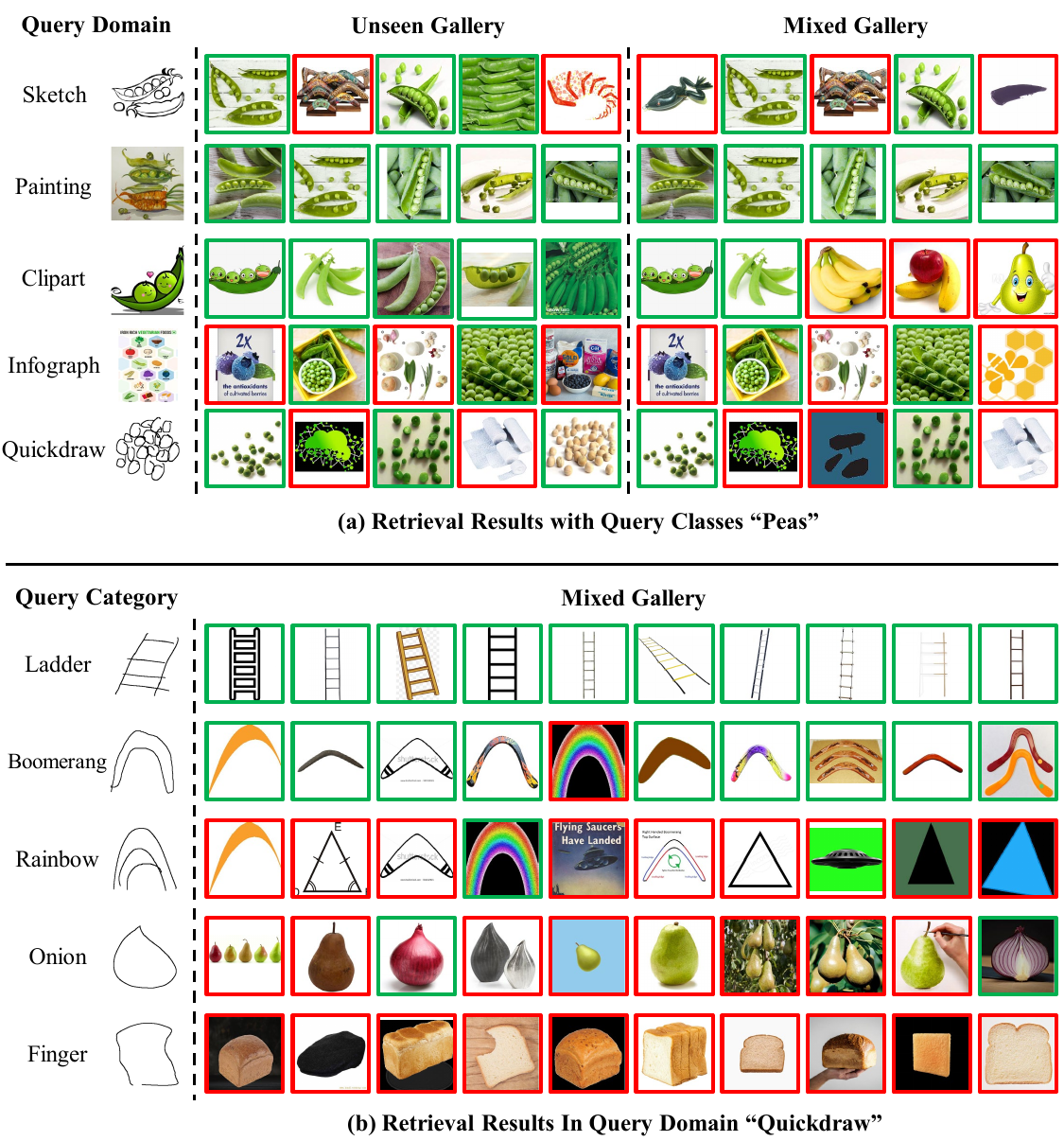}
    \caption{Retrieval Results under UCDR protocols on DomainNet. \textbf{(a)} displays the retrieval results of ``Peas" by the query from an unseen query domain. \textbf{(b)} shows the retrieval results of a few queries from Quickdraw. True positives and false positives are shown with green and red borders, respectively.}
    \label{fig:qualitative}
    \vspace{-1.5em}
\end{figure}

\vspace{1.5em}
\noindent \textbf{Impact of two prompts length.}
We further conduct experiments to analyze the impact of two key prompt lengths in ProS. \ie, text prompts and Content-aware Dynamic Prompt (CaDP). \cref{fig:prompt length}(a) shows that when text prompts length increases, the performance improves to its peak when the length is 16 and then gradually decreases. Therefore, we set the text prompt
length to 16 for all datasets. In \cref{fig:prompt length}(b), we find that ProS achieves the state-of-the-art with only 1-1 CaDP. We speculate that this might be consistent with Prompt Units, which also use a single prompt to represent one domain or category.

\subsection{Qualitative Analysis}

We visualize the feature space construct by CLIP-Full, VPT and ProS for 10 randomly selected unseen categories from \textit{Infograph} Query and \textit{Real} Gallery sets, as shown in \cref{fig:visualization}. 
To further compare the feature space constructed by three methods, we evaluate the inter-class distinctiveness and intra-class compactness of feature space by the metric $\sigma=\frac{\max\mathcal{D_{\mathit{intra}}}}{\min\mathcal{D_{\mathit{inter}}}}$ from \cite{LBHash} where $\mathcal{{D_\mathit{intra}}}$ means intra-class distance and $\mathcal{D_{\mathit{inter}}}$ means inter-class distance. We can observe that: \textbf{First,} the features obtained by fine-tuned CLIP are not sufficiently separated and are not effectively aligned gallery and query domain, while VPT can partially alleviate this issue. \textbf{Second,} our ProS can extract more clustered features of the same classes compared with both fine-tuned CLIP and VPT, demonstrating the superiority of our approach. \textbf{Third,} comparing the extracted features using metrics from \cite{LBHash}, we draw the consistent conclusion that CLIP $<$ VPT $\ll$ ProS.


To further demonstrate the effectiveness of our ProS, we visualize a few retrieval results produced by ProS. \cref{fig:qualitative} (a) shows the top five retrieved candidates of peas with query images from different domains, while \cref{fig:qualitative} (b) shows the top ten retrieved candidates with queries from Quickdraw. From the figure, we can see that Sketch and Quickdraw images lack enough information such as color, texture, and details while Inforgraph has an object co-occurrence problem. They all obtain worse performance compared with other domains. This phenomenon will be exacerbated when the gallery contains both seen and unseen classes due to the interference caused by seen classes. Towards this phenomenon, we can conclude that: due to the quality issues of the dataset, the performance of our APL in certain domains may be relatively poor. 
\section{Conclusion}
In this paper, to the best of our knowledge, we take the first attempt to deploy CLIP for Universal Cross-Domain Retrieval (UCDR) with a prompt tuning paradigm and propose a more generalized prompt method named Prompt-to-Simulate (ProS). Specifically, ProS can dynamically fit unknown domain and category distribution with guidance of source knowledge by a two-stage training paradigm following a mask-and-align objective. With the above approach, our model obtains quite strong performance under UCDR compared to the state-of-the-art.

{
    \small
    \bibliographystyle{ieeenat_fullname}
    \bibliography{main}
}


\end{document}